\documentclass[11pt,a4paper]{article}
\usepackage[hyperref]{emnlp-ijcnlp-2019}
\usepackage{times}
\usepackage{latexsym}

\usepackage{url}
\usepackage{graphicx}
\usepackage{subcaption}
\usepackage{dblfloatfix}
\usepackage{bm}
\usepackage{amsfonts}
\usepackage{amsmath}
\usepackage{multirow}
\usepackage[ruled, vlined]{algorithm2e}
\usepackage{soul}
\usepackage{array}
\newcolumntype{P}[1]{>{\centering\arraybackslash}p{#1}}
\aclfinalcopy % Uncomment this line for the final submission

\newcommand{\fromH}[1]{\textcolor{blue}{#1}}
\newcommand{\xd}[1]{\textcolor{brown}{#1}}

\newcommand{\reds}{\texttt{REDS2}}
\newcommand{\bag}{1-hop DS bag}
\newcommand{\tbag}{2-hop DS bag}
\newcommand{\nop}[1]{}

\title{Leveraging \textit{2-hop} Distant Supervision from Table Entity Pairs for Relation Extraction}

\author{Xiang Deng \\
  The Ohio State University \\
  {\tt deng.595@buckeyemail.osu.edu} \\\And
  Huan Sun \\
  The Ohio State University \\
  {\tt sun.397@osu.edu} \\}

\date{}

\begin{document}
\maketitle
\begin{abstract}
{Distant supervision (DS) has been widely used to automatically construct (noisy) labeled data for relation extraction (RE)\nop{from unstructured text}. Given two entities, distant supervision exploits sentences that directly mention them for predicting their semantic relation. We refer to this strategy as \text{1-hop} DS, which unfortunately may not work well for long-tail entities with few supporting sentences. In this paper, we introduce a new strategy named \textit{2-hop} DS to enhance distantly supervised RE, based on the observation that there exist a large number of relational tables on the Web which contain entity pairs that share common relations. We refer to such entity pairs as \textit{anchors} for each other, and collect all sentences that mention the anchor entity pairs of a given target entity pair to help relation prediction. We develop a new neural RE method \reds\ in the multi-instance learning paradigm, which adopts a hierarchical model structure \nop{and a dynamic bag aggregation mechanism}to fuse information respectively from 1-hop DS and 2-hop DS. Extensive experimental results on a benchmark dataset show that \reds\ can consistently outperform various baselines across different settings by a substantial margin.\footnote{Our source code and datasets are at \url{https://github.com/sunlab-osu/REDS2}.} \nop{We also conduct case studies to show that our model is capable of finding new knowledge by incorporating Web tables.}}
%Distant supervised relation extraction completes Knowledge Base (KB) by extracting new relation facts from unstructured texts. However, most existing methods only use sentences that can directly align with the target entity pair and perform poorly with long-tail entities. Meanwhile, tables and structured lists on the Web are a potential source of knowledge, which is ignored by current RE systems. In this paper, we introduce XXXX, a distant supervised neural relation extraction method that leverages entity pair correlations found in tables for improved relation extraction. It uses entity pairs co-occurred in tables to augment each other and fuses the extra information from tables dynamically. The experimental results on benchmark dataset show significant and consistent improvements on relation extraction as compared with baselines. We also conduct experiments to show that our model is capable of finding new knowledge by incorporating Web tables.
\end{abstract}
\section{Introduction}
Relation extraction (RE) aims to extract semantic relations between two entities from unstructured text and is an important task in natural language processing (NLP). Formally, given an entity pair $(e_1,e_2)$ from a knowledge base (KB) and a sentence (instance) that mentions them, RE tries to predict if a relation $r$ from a predefined relation set exists between $e_1$ and $e_2$. A special relation \textit{NA} is used if none of the predefined relations holds.  

\begin{figure}[t!]
    \centering
    \includegraphics[width=\linewidth]{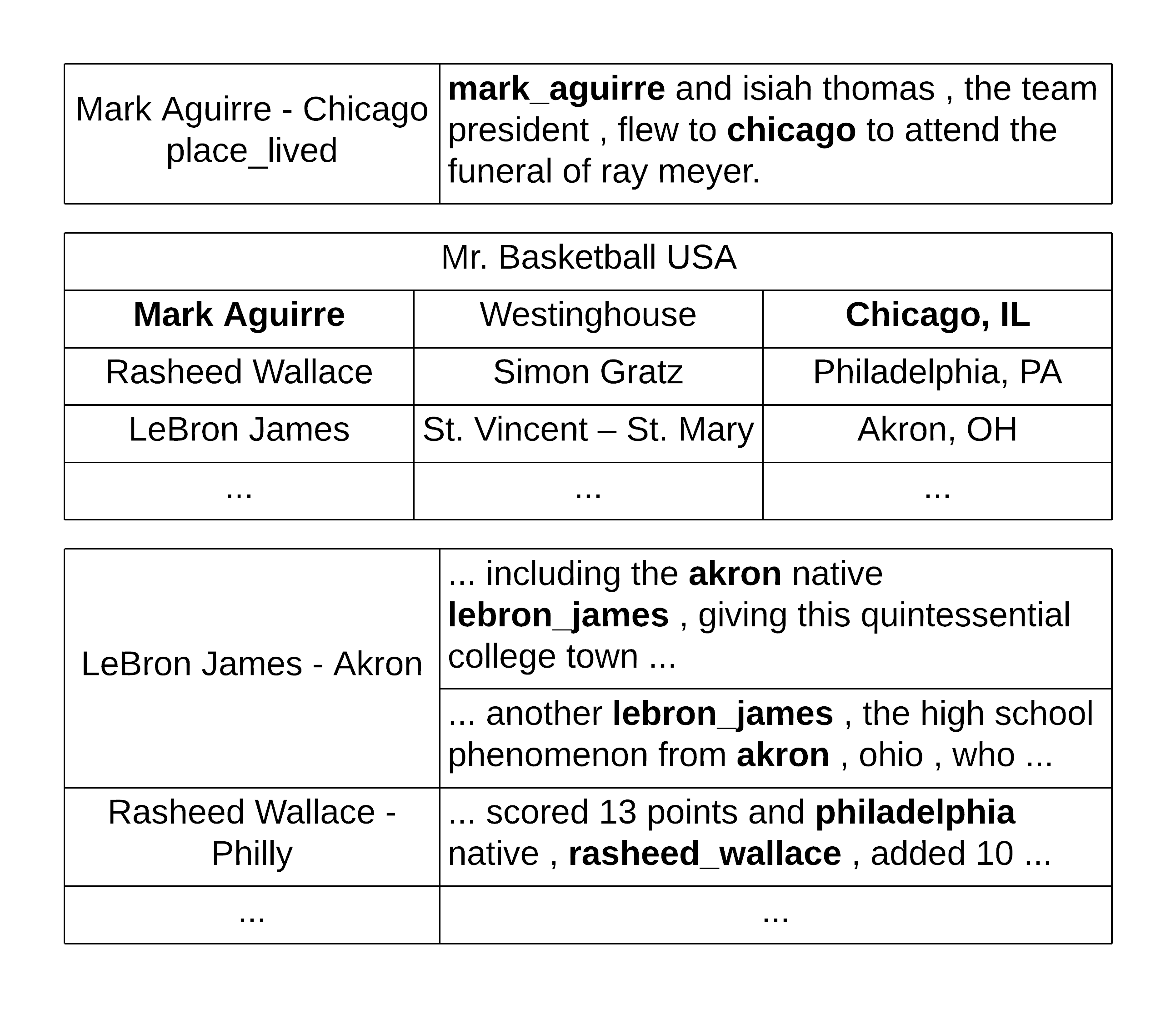}
    %\caption{\nop{\fromH{It is better to show some concrete examples, e.g., a table with two rows (Actor 1, award 1), (Actor 2, award 2) and other rows are "..." and then show some of the sentences from 1-hop DS for the first pair and some of the sentences for the second pair. This example is to show your intuition/observation in the current 3rd paragraph...  }} \nop{An example of web table listing winners of Mr. Basketball USA, and the sentences associated with the linked entity pairs. The relations are $\textcircled{\scriptsize{1}}\langle$ award.winner$\rangle$, $\textcircled{\scriptsize{2}}\langle$education.highschool$\rangle$ and $\textcircled{\scriptsize{3}}\langle$people.person.place\_of\_birth$\rangle$}An example of 2-hop DS. In the top is the target entity pair, the relation is \textit{place\_of\_birth} but we cannot infer it from the corresponding sentence. In the middle is a table get from Wikipedia page of \textit{Mr. Basketball USA}, where we can find several anchors for the target entity pair. In the bottom are sentences associated to the anchors, which indicate the relation more clearly.}
    \caption{Illustration of \textit{2-hop} distant supervision. The top panel shows a target entity pair, one sentence that mentions it, and the relation under study which cannot be inferred from the sentence. The middle gives part of a table from Wikipedia page ``\textit{Mr. Basketball USA}'', where we can extract \textit{anchors} for the target entity pair. The bottom shows some sentences that are associated with the anchors, which more clearly indicate the underinvestigated relation and can be utilized to extract relations between the target entity pair.}
    \label{fig:table_example}
\end{figure}

Given that it is costly to construct large-scale labeled instances for RE,  distant supervision (DS) has been a popular strategy to automatically construct (noisy) training data. It assumes that if two entities hold a relation in a KB, all sentences mentioning them express the same relation. Noticing that the DS assumption does not always hold and has the wrong labeling problem, many efforts including \citep{riedel2010modeling,hoffmann2011knowledge,surdeanu2012multi} have adopted the multi-instance learning paradigm to tackle the challenge, and more recently, neural models with attention mechanism have been proposed to de-emphasize the noisy instances\nop{focus more on sentences that are more likely to express a certain relation} \citep{lin2016neural,ji2017distant,han2018hierarchicalRE}. Such models tend to work well when there are a large number of sentences talking about the target entity pair \citep{lin2016neural}. \nop{The superiority of the prevalent attention-based multi-instance learning frameworks lies in the entity pairs having multiple sentences \citep{lin2016neural}.} However, we observe that there can be a large portion of entity pairs that have very few supporting sentences (e.g., nearly 75\% of entity pairs in the \citet{riedel2010modeling} dataset only have one single sentence mentioning them), which makes distantly supervised RE even more challenging.%How to improve performance on these long-tail entities is still an open question.}

%\fromH{In this paragraph, we will clarify how we obtain more distant supervision via table entity pairs, why it would make sense, some numbers showing the coverage, and we name it 2-hop distant superision. move the following part to related work.}

The conventional distant supervision strategy only exploits instances that directly mention a target entity pair, and because of this, we refer to it as \textit{1-hop} distant supervision. On the other hand, there are a large number of Web tables that contain relational facts about entities \citep{cafarella2008webtables,venetis2011recovering, wang2012understanding}. Owing to the semi-structured nature of tables, we can extract from them sets of entity pairs that share common relations, and sentences mentioning these entity pairs often have similar semantic meanings. Under this observation, we introduce a new strategy named \textit{2-hop} distant supervision: We define entity pairs that potentially have the same relation with a given target entity pair as \textit{anchors}, which can be found through Web tables, and aim to fully exploit the sentences that mention those anchor entity pairs to augment RE for the target entity pair. Figure \ref{fig:table_example} illustrates the 2-hop DS strategy.%{Given a table, we can extract sets of entity pairs that share common relations, and sentences mentioning these entity pairs often have similar semantic meanings. Under this observation, we introduce a new strategy named \textit{2-hop} distant supervision: we define entity pairs that potentially have the same relation with a given target entity pair as \textit{anchors}, which can be found through Web tables, and aim to fully exploit the sentences that mention those anchor entity pairs to augment relation prediction for the target entiy pair. Figure \ref{fig:table_example} illustrates the 2-hop DS strategy.}
\nop{We notice that entities pairs under the same columns generally share common relations, and sentences mentioning these entities often have similar semantic meanings. Under this observation, we introduce a new strategy named \textit{2-hop} distant supervision: We define entity pairs that co-occur with a given target entity pair in the same table columns as \textit{anchors}, and aim to fully exploit the sentences that mention those anchor entity pairs to augment relation prediction for the target entiy pair. Figure \ref{fig:table_example} illustrates the 2-hop DS strategy.}

The intuition behind 2-hop DS is if the target entity pair holds a certain relation, one of its anchors is likely to have that relation too and at least one sentence mentioning the anchors should express the relation. Despite being noisy, the 2-hop DS can provide extra, informative supporting sentences for the target entity pair. One straightforward approach is to merge the two bags of sentences respectively derived from 1-hop and 2-hop DS as one single set and apply existing multi-instance learning models. However, \nop{the 2-hop DS strategy inevitably aggravates the wrong labeling problem that already exists in 1-hop DS.}the 2-hop DS strategy also has the wrong labeling problem that already exists in 1-hop DS. Simply mixing the two sets of sentences together may mislead the prediction, especially when there is a great disparity in their size. In this paper, we propose \reds\footnote{stands for relation extraction with 2-hop DS.}, a new neural relation extraction method in the multi-instance learning paradigm, and design a hierarchical model structure to fuse information from 1-hop and 2-hop DS. We evaluate \reds\ on a widely used benchmark dataset and show that it consistently outperforms various baseline models by a large margin. %Blindly,across different settings 

We summarize our contributions as three-fold:
\begin{itemize}
    \item We introduce 2-hop distant supervision as an extension to the conventional distant supervision, and leverage entity pairs in Web tables as anchors to find additional supporting sentences to further improve RE.
    \item We propose \reds, a new neural relation extraction method based on 2-hop DS and has achieved new state-of-the-art performance in the benchmark dataset \cite{riedel2010modeling}.
    \item We release both our source code and an augmented benchmark dataset that has entity pairs aligned with those in Web tables, to facilitate future work.
\end{itemize}
%that aligns entity pairs in the benchmark dataset with those in web tables

\begin{figure*}[htp!]
    \centering
    \includegraphics[width=\textwidth]{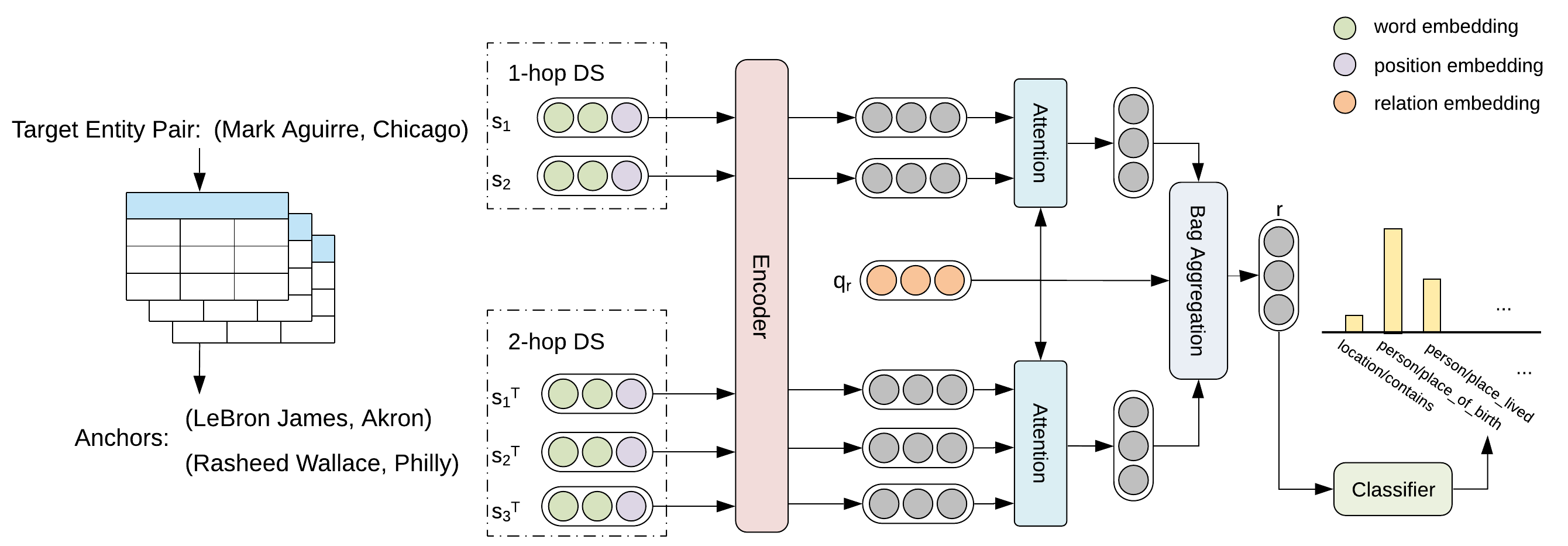}
    \caption{Overview of our method \reds. \reds\ first obtains \textit{anchors} of the target entity pair and constructs a 2-hop DS bag. Sentences in the 1-hop and 2-hop DS bag are individually encoded with a PCNN sentence encoder \citep{zeng2015distant}. We then use selective attention and bag aggregation to get the final representation, based on which a classifier predicts scores for each candidate relation.}%REDS2 first applies 2-hop distant supervision to the target entity pair and construct a 2-hop DS instance bag. Sentences in 1-hop DS bag and 2-hop DS bag are then encoded with PCNN sentence encoder. We then use selective attention and dynamic bag aggregation to get final representation. In the end, a classifier is used to predict scores for each relation based on the final representation.
    \label{fig:framework}
\end{figure*}

\section{Related Work}
\noindent \textbf{Distant Supervision.} One main drawback of traditional supervised relation extraction models \citep{zelenko2003kernel,mooney2006subsequence} is they require adequate amounts of annotated training data, which is time consuming and labor intensive. To address this issue, \citet{mintz2009distant} proposes distant supervision (DS) to automatically label data by aligning plain text with Freebase. However, DS inevitably accompanies with the wrong labeling problem. To alleviate the noise brought by DS, \citet{riedel2010modeling} and \citet{hoffmann2011knowledge} introduce multi-instance learning mechanism, which is originally used to combat the problem of ambiguously-labeled training data when predicting the activity of different drugs \citep{dietterich1997solving}. 

\noindent \textbf{Neural Relation Extraction.} Early stage relation extraction (RE) methods use features extracted by NLP tools and strongly rely on the quality of features. Due to the recent success of neural models in different NLP tasks, many researchers have investigated the possibility of using neural networks to build end-to-end relation extraction models. \citet{zeng2014relation} uses convolutional neural network (CNN) to encode sentences, which is further improved through piecewise-pooling \citep{zeng2015distant}. \citet{adel2017global} and \citet{gupta2016table} use neural networks for joint entity and relation extraction. More advanced network architectures like Tree-LSTM \citep{miwa2016end} and Graph Convolution Network \citep{vashishth2018reside} are also adopted to learn better representations by using syntactic features like dependency trees. Most recent models also incorporate neural attention technology \citep{lin2016neural} as an improvement to at-least-one multi-instance learning \citep{zeng2015distant}. \citet{han2018hierarchicalRE} further develops a hierarchical attention scheme to utilize the relation correlations and help predictions for long-tail relations.

\noindent\textbf{Web Table Understanding.} Aside from plain texts, there are large amounts of factual knowledge in the Web expressed in hundreds of millions of tables and other structured lists \citep{cafarella2008webtables,venetis2011recovering}, which have not been fully explored yet. Table understanding tries to match tables to KB and parse the schemas. Existing methods for table understanding mainly fall into two categories. One is based on local evidence \citep{venetis2011recovering,munoz2014using,ritze2015matching}. Given one table, the main idea is to first link cells to entities in KB. We can then use existing relations between linked entities to infer relations between columns and extract new facts by generalizing to all rows. However, this method requires a high overlap between table and KB, which is hampered by KB incompleteness. The other approach tries to leverage features extracted from the table header and column names \citep{ritze2017matching,cannaviccio2018leveraging}. Unfortunately, a large portion of Web tables miss such metadata or contain limited information, and the second approach will fail in such cases. Although the focus of this paper is the RE task, we believe the idea of connecting Web tables and plain texts using DS can potentially benefit table understanding as well.
\section{Methodology}
%\fromH{This section can be polished according to the definition of terms in intro: anchor entity pairs, and the intuition why table helps.}

Given a set of sentences $\mathrm{S} = \{s_1,s_2,...\}$ and a target entity pair $(h, t)$, we will leverage the directly associated sentence bag $S_{h,t} \subseteq \mathrm{S}$ by 1-hop distant supervision (1-hop DS bag), and the table expanded sentence bag $S_{h,t}^T \subseteq \mathrm{S}$ by 2-hop distant supervision (2-hop DS bag), for relation extraction. $S_{h,t}$ contains all instances mentioning both $h$ and $t$, while $S_{h,t}^T$ is obtained indirectly through the anchors of $(h, t)$ found in Web tables. Following previous work~\citep{riedel2010modeling,hoffmann2011knowledge}, we adopt the multi-instance learning paradigm to measure the probability of $(h, t)$ having relation $r$. 

%In this section we introduce the three main components of our model:
Figure \ref{fig:framework} gives an overview of our framework with three major components:
\begin{itemize}
\item \textbf{Table-aided Instance Expansion:}
    Given a target entity pair $(h,t)$, we find its \textit{anchor} entity pairs $\{(h_1,t_1),(h_2,t_2),...\}$ through Web tables. We define an anchor entity pair as two entities co-occurring with $(h,t)$ in some table columns at least once. $S_{h,t}^T=S_{h_1,t_1}\cup S_{h_2,t_2} \cup ... $ is then exploited to augment the directly associated bag $S_{h,t}$.
    
\item \textbf{Sentence Encoding:}
    For each sentence $s$ in bag $S_{h,t}$ or $S_{h,t}^T$, a sentence encoder is used to obtain its semantic representation $\mathbf{s}$.

\item \textbf{Hierarchical Bag Aggregation:} 
    Once the embedding of each sentence is learned, we first use a sentence-level attention mechanism to get bag representation $\mathbf{h}$ and $\mathbf{h}^T$, and then aggregate them for final relation prediction. \nop{They are then weighted and summed to obtain the final representation, from which we make our prediction.}\nop{and then \textit{dynamically} aggregate \fromH{Is "dynamic" here the appropriate word? Check copy mechanism?} them together for final prediction.}
\end{itemize}
\subsection{Table-aided Instance Expansion} \label{sec:ins_expand}
Now we introduce how to construct the table expanded sentence bag $S^T_{h,t}$ for a given target entity pair $(h,t)$ by 2-hop distant supervision. %through Web tables.
%\fromH{This section needs more efforts to polish}
\subsubsection{Web Tables}
Web tables have been found to contain rich facts of entities and relations. \nop{A large number of Web tables are entity-focused, constructed by a group of entities focusing on a certain topic. Entities in the same row usually have certain relations between each other, and this relation holds across the rows.We use tables from Wikipedia \citep{bhagavatula2015tabel} as our table corpus $T=\{\mathrm{t_1},\mathrm{t_2},...\}$. A typical entity-focused Wikipedia table $\mathrm{t}$ has the following elements:}It is estimated that out of a total of 14.1 billion tables on the Web, 154 million tables contain relational data \citep{cafarella2008webtables} and Wikipedia alone is the source of nearly 1.6 million relational tables \citep{bhagavatula2015tabel}. Columns of a Wikipedia table can be classified into one of the following data types: `empty', `named entity', `number', `date expression', `long text' and `other' \citep{zhang2017effective}. Here we only focus on named entity columns (NE-columns) and the Wikipedia page title, which can be easily linked to KB entities. These entities can be further categorized as:%A typical relational table in Wikipedia is constructed by a set of entities focusing on a certain topic which can be further categorized as follows:

{\bf A topic entity $\bm{e^t}$} that the table is centered around. We refer to the Wikipedia article where the table is found and take the entity it describes as $e^t$.

{\bf Subject entities $\bm{E^s=\{e^s_1,e^s_2,...\}}$} that can act as primary keys of the table. Following previous work on Web table analysis \citep{venetis2011recovering}, we select the {leftmost NE-column} as subject column and its entities as $E^s$. %\fromH{Question here: when people go to the "Mr. Basketball USA" page and look at the table, they will see the year column as distinct? right? Should we see linked to KB? besides, they will see two LeBron James... }

{\bf Body entities $\bm{E=\{e_{1,1},e_{1,2},...\}}$} that compose the rest of the table. All entities in non-subject NE-columns are considered as $E$.%\fromH{the description in this paragraph seems less precise.}

\subsubsection{2-hop Distant Supervision} 
In the conventional distant supervision setting, each entity pair $(h,t)$ is associated with a bag of sentences $S_{h,t}$ that directly mention $h$ and $t$. The intuition behind 2-hop distant supervision is, if $(h_i,t_i)$ and $(h_j,t_j)$ potentially hold the same relation, we can treat them as \textbf{anchor entity pairs} for each other, and then use the \bag\ $S_{h_j,t_j}$ to help with the prediction for $(h_i,t_i)$ and vice versa. In this paper, we extract anchor entity pairs with the help of Web tables.

We notice that owing to the semi-structured nature of tables, (1) subject entities can usually be connected with the topic entity by the same relation. (2) Non-subject columns of a table usually have binary relationships to or are properties of the subject column. Body entities in the same column share common relations with their corresponding subject entities. For example, in Figure \ref{fig:table_example}, the topic entity is ``\textit{Mr. Basketball USA}''; column 1 is the subject column and contains a list of winners of ``\textit{Mr. Basketball USA}''; column 2 and column 3 are high school and city of the subject entity.
\nop{Intuitively, subject entities are connected to the topic entity by the same relation, and body entities in the same column share common relations with their subject entities.}
\nop{entities in the same row of a table usually have a certain relation, and this relation holds across all rows. If $(h_i,t_i)$ and $(h_j,t_j)$ co-occur in a pair of table columns, we treat them as \textbf{anchor entity pairs} for each other, and use $S_{h_j,t_j}$ to help with the prediction for $(h_i,t_i)$ and vice versa.
\noindent \textbf{anchor entity pairs.} Intuitively, body entities in the same column share common relation with their subject entities.}

Formally, we consider two entity pairs $(h_i,t_i)$ and $(h_j,t_j)$ as anchored if there exists a Web table such that either criterion below is met: %via a Web table if they meet either criterion below:
\begin{itemize}
    \item $h_i=h_j=e^t$ and $t_i,\,t_j \in E^s$.
    \item $h_i \in E^s$ or $t_i \in E^s$, $(h_i, h_j)$ is in the same column (and so is $(t_i, t_j)$), and $(h_i, t_i)$ is in the same row (and so is $(h_j, t_j)$) \nop{$h_ih_j)$/$(t_i,t_j)$ are in the same column and $(h_i,t_i)$/$(h_j,t_j)$ are in the same row.}
\end{itemize}

%\fromH{I changed the logic flow because we have never mentioned the topic/subject entity pairs before.}\xd{Xiang: Now I first give difinition of 2-hop DS, then explain how we extract anchor entity pairs from tables. Is this ok?}
\nop{In addition, we notice that . Hence we also consider entity pairs $(h_i,t_i)$ and $(h_j,t_j)$ as anchored if $h_i=h_j=e^t$ and $t_i,t_j \in E^s$. }

The \tbag\ $S^T_{h,t}$ is then constructed as the union of $S_{h_i,t_i}$'s, where $(h_i,t_i)$ is an anchor entity pair of $(h,t)$.
%$\{S_{h_i,t_i},...\}$ if $\exists \mathrm{t} \in T$, $(h,t),(h_i,t_i)$ are anchor entity pairs in $\mathrm{t}$.

\subsection{Sentence Encoding}
Given a sentence $s$ consisting of $n$ words $s = \{w_1,w_2,..., w_n\}$, we use a neural network with an embedding layer and an encoding layer to obtain its low-dimensional vector representation.
\subsubsection{Embedding Layer}
Each token is first fed into an embedding layer to embed both semantic and positional information.

{\bf Word Embedding} maps words to vectors of real numbers which preserve syntactic and semantic information of words. Here we get a vector representation $\mathbf{w}_i \in \mathbb{R}^{k_w}$ for each word from a pre-trained word embedding matrix.

{\bf Position Embedding} was proposed by \citet{zeng2014relation}. Position embedding is used to embed the positional information of each word relative to the head and tail mention. A position embedding matrix is learned in training to compute position representation $\mathbf{p}_i \in \mathbb{R}^{k_p \times 2}$. %\fromH{not very familiar with the position embedding, why *2?}.

Finally, we concatenate the word representation $\mathbf{w}_i$ and position representation $\mathbf{p}_i$ to build the input representation $\mathbf{x}_i \in \mathbb{R}^{k_i}$ (where $k_i=k_w+k_p \times 2$) for each word $w_i$.
\subsubsection{Encoding Layer}
A sequence of input representations $\mathbf{x}=\{\mathbf{x}_1,\mathbf{x}_2,...\}$ with a variable length is then fed through the encoding layer and converted to a fixed-sized sentence representation $\mathbf{s} \in \mathbb{R}^{k_h}$. There are many existing neural architectures that can serve as the encoding layer, such as CNN \citep{zeng2014relation}, PCNN \citep{zeng2015distant} and LSTM-RNN \citep{miwa2016end}. We simply adopt PCNN here, which has been shown very powerful and efficient by a number of previous RE works. 

PCNN is an extension to CNN, which first slides a convolution kernel with a window size $m$ over the input sequence to get the hidden vectors: %\fromH{If later we run out of space, make equation (1) and (3) inline or remove part of this section?}:
\begin{equation}
    \mathbf{h}_i = \mathrm{CNN}(\mathbf{x}_{i-\frac{m-1}{2}:i+\frac{m-1}{2}}),
\end{equation}
A piecewise max-pooling is then applied over the hidden vectors:
\begin{align}
    [\mathbf{s}^{(1)}]_j&=\underset{1 \leq i \leq i_1}{\mathrm{max}}\{[\mathbf{h}_i]_j\},\nonumber\\
    [\mathbf{s}^{(2)}]_j&=\underset{i_i+1 \leq i \leq i_2}{\mathrm{max}}\{[\mathbf{h}_i]_j\},\\
    [\mathbf{s}^{(3)}]_j&=\underset{i_2+1 \leq i \leq n}{\mathrm{max}}\{[\mathbf{h}_i]_j\},\nonumber
\end{align}
where $i_1$ and $i_2$ are head and tail positions. The final sentence representation $\mathbf{s}$ is composed by concatenating these three pooling results $
    \mathbf{s} = [\mathbf{s}^{(1)};\mathbf{s}^{(2)};\mathbf{s}^{(3)}].
$
\subsection{Hierarchical Bag Aggregation}
After we get sentence representations $\{\mathbf{s}_1,\mathbf{s}_2,...\}$ and $\{\mathbf{s}_1^T,\mathbf{s}_2^T,...\}$ for $S$ and $S^T$, to fuse key information from these two bags, we adopt a hierarchical aggregation design to obtain the final representation $\mathbf{r}$ for prediction. We first get bag representation $\mathbf{h}$ and $\mathbf{h}^T$ using a sentence-level selective attention, and then employ a bag-level aggregation to compute $\mathbf{r}$.
\subsubsection{Sentence-level Selective Attention}
Since\nop{sentences in $S_{h,t}$ and $S_{h,t}^T$ are all collected by distant supervision,} the wrong labeling problem inevitably  exists in both 1-hop and 2-hop distant supervision, here we use selective attention to assign different weights to different sentences given relation $r$ and de-emphasize the noisy sentences. The attention is caculated as follows:
\begin{align}
    e_i &= \mathbf{q}_r^{\mathsf{T}}\mathbf{s}_i, \nonumber\\
    \alpha_i &= \frac{\mathrm{exp}(e_i)}{\sum_{j=1}^{n}\mathrm{exp}(e_j)}, \\
    \mathbf{h} &= \sum_{i=1}^{n} \alpha_i\mathbf{s}_i, \nonumber
\end{align}
where $\mathbf{q}_r$ is a query vector assigned to relation $r$. $\mathbf{h}$ and $\mathbf{h}^T$ are computed respectively for the two bags $S$ and $S^T$.
%\subsubsection{Dynamic Bag Aggregation}
\subsubsection{Bag-level Aggregation}
Since \tbag\ $S^T$ is collected indirectly through anchor entity pairs in Web tables, despite that it brings abundant information, it also contains a massive amount of noise. Thus treating $S^T$ equally as $S$ may mislead the prediction, especially when their sizes are extremely imbalanced. 

\nop{Here we utlize information from $\mathbf{h}$, $\mathbf{h}^T$ and $\mathbf{q}_r$ to balance between $S_{h,t}^T$ and $S_{h,t}$, \fromH{It could be better to explain why we use these to predict the weight.. and why it is a ``dynamic" weight..} and calculate the weight $\beta$ for 1-hop DS as:}
To automatically decide how to balance between $S$ and $S^T$, we utilize information from $\mathbf{h}$, $\mathbf{h}^T$ and $\mathbf{q}_r$ to predict a weight $\beta$:
\begin{equation}
\beta = \sigma(\mathbf{W}[\mathbf{h};\mathbf{h}^T;\mathbf{q}_r] + \mathbf{b}),
\end{equation}
where vector $\mathbf{W}$ and scalar $\mathbf{b}$ are learnable variables and $\sigma$ is the sigmoid function. Next, $\beta$ is used as a weight to fuse information from 1-hop DS and 2-hop DS, determined by $S$ and $S^T$ of the current target entity pair and relation $r$. We then obtain the final representation $\mathbf{r}$ as:
\begin{equation}
\mathbf{r} = \beta \mathbf{h} + (1-\beta)\mathbf{h}^T,
\label{eq:dynamic_agg}
\end{equation}

Finally, we define the conditional probability $P(r|S,S^T,\theta)$ as follows,
\begin{equation}
    P(r|S,S^T,\theta) = \frac{\mathrm{exp}(\mathbf{o}_r)}{\sum^{n_r}_{k=1}\mathrm{exp}(\mathbf{o}_k)}，
\end{equation}
where $\mathbf{o}$ is the score vector for current target entity pair having each relation\nop{each relation $r$ based on $\mathbf{r}$},
\begin{equation}
    \mathbf{o}=\mathbf{Mr}+\mathbf{d},
\end{equation}
here $\mathbf{M}$ is the representation matrix of relations, which shares weights with $\mathbf{q}_r$'s. $\mathbf{d}$ is a learnable bias term. %\fromH{which is composed of $\mathbf{q}_r$'s?}.

\subsection{Optimization}% and Implementation Details

%Here we introduce the optimization and implementation details of our model.
We adopt the cross-entropy loss as the training objevtive function. Given a set of target entity pairs with relations $\pi = \{(h_1,t_1,r_1),(h_2,t_2,r_2),...\}$, we define the loss function as follows:
\begin{equation}
    J(\theta) = -\frac{1}{|\pi|}\sum_{i=1}^{|\pi|}\mathrm{log}P(r_i|S_{h_i,t_i},S_{h_i,t_i}^T,\theta).
\end{equation}
All models are trained with stochastic gradient descent (SGD) to minimize the objective function. The same sentence encoder is used to encode $S$ and $S^T$.

\section{Experiments}
\subsection{Datasets and Evaluation}

We evaluate our model on the New York Times (NYT) dataset developed by \citet{riedel2010modeling}, which is widely used in recent works. The dataset has 53 relations including a special relation $\mathrm{NA}$ which indicates none of the other 52 relations exists between the head and tail entity.

We use the WikiTable corpus collected by \citet{bhagavatula2015tabel} as our table source. It originally contains around 1.65M tables extracted from Wikipedia pages. Since the NYT dataset is already linked to Freebase, we perform entity linking on the table cells and the Wikipedia page titles using existing mapping from Wikipedia URL to Freebase MID (Machine Identifier). We then align the table corpus with NYT and construct $S^T$ for entity pairs as detailed in section \ref{sec:ins_expand}. \textit{For both training and testing, we only use entity pairs and sentences in the original NYT training data for table-aided instance expansion.} We set the max size of $S^T$ as 300, and randomly sample 300 sentences if $|S^T|>300$. Statistics of our final dataset is summarized in Table \ref{tab:datastat}. One can see that 38.18\% and 46.79\% of relational facts (i.e., entity pairs holding non-NA relations) respectively in the training and testing set can potentially benefit from leveraging 2-hop DS. %\fromH{A bit confused by this statement; see Table 1}%benefited from 2-DS. 

Following prior work \citep{mintz2009distant}, we use the testing set for held-out evaluation, and evaluate models by comparing the predicted relational facts with those in Freebase. For evaluation, we rank the extracted relational facts based on model confidence and plot precision-recall curves. In addition, we also show the area under the curve (AUC) and precision values at specific recall rates to conduct a more \nop{quantitative}comprehensive comparison.
\begin{table}[t!]
    \centering
    \resizebox{\linewidth}{!}{
    \begin{tabular}{|c|cc|cc|}
    \cline{2-5}
        \multicolumn{1}{c|}{} &  \multicolumn{2}{c|}{Train}&\multicolumn{2}{c|}{Test}\\
       \multicolumn{1}{c|}{}  & Overall\nop{All} & Non-NA & Overall\nop{All} & Non-NA \\
    \hline
    \# Entity Pairs &291699&18144&96678&1761\\
    %\# Table Entity Pairs &17565&6928&4832&824\\
    \# Entity Pairs with $|S^T|>0$ &17565&6928&4832&824\\
    \% Entity Pairs with $|S^T|>0$ &6.02&\textbf{38.18}&5.00&\textbf{46.79}\\
    mean $|S|$ &1.69&5.24&1.62&2.78\\
    mean $|S^T|$ &147.65&190.61&131.65&217.23\\
    \hline
    \end{tabular}}
    \caption{Dataset statistics. We show statistics of entity pairs that hold non-NA relations separately from overall, as they are important relational facts to discover. Among non-NA entity pairs, 38.18\% in training and 46.79\% in testing have nonempty $S^T$, which respectively have 190.61 and 217.23 sentences on average. \nop{Table entity pairs are enity pairs with nonempty $S^T$. Mean $|S^T|$ is calculated based on table entity pairs.}}
    \label{tab:datastat}
\end{table}
%\subsection{Experimental Settings}

%\subsection{Overall Evaluation Results}
\subsection{Baselines}
%To evaluate the performance of our proposed method \reds, 
We compare \reds\ with the following baselines:

\noindent\textbf{PCNN+ATT} \citep{lin2016neural}. This model uses a PCNN encoder combined with selective attention over sentences.  Since this is also the base block of our model, we also refer to it as BASE in this paper.

\noindent\textbf{PCNN+HATT} \citep{han2018hierarchicalRE}. This is another PCNN based relation extraction model, where the authors use hierarchical attention to model the semantic correlations among relations.

\noindent\textbf{RESIDE} \citep{vashishth2018reside}. It uses Graph Convolutional Networks (GCN) for sentence encoding, and also leverages relevant side information like relation alias and entity type.

Results of PCNN+HATT and RESIDE are directly taken from the code repositories released by the authors. For PCNN+ATT, we report results obtained by our reproduced model, which are close to those shown in \citep{lin2016neural}. To simply verify the effectiveness of adding extra supporting sentences from 2-hop DS, we also compare the following vanilla method with PCNN+ATT:\nop{To further verify the effectiveness of our hierarchical bag aggregation design, we also compare with the following vanilla method to utilize 2-hop DS}

\noindent\textbf{BASE+MERGE}. For each target entity pair $(h, t)$, we simply merge $S$ and $S^T$ as one sentence bag, and apply the trained PCNN+ATT (or, BASE) model. %This method simply merges $S_{h,t}$ with $S_{h,t}^T$ and applies the pre-trained BASE model. %\fromH{No re-training?}

\subsection{Implementation Details}
\begin{table}
    \centering
    \begin{tabular}{c|c}
    \hline
    Window size $m$ &  3\\
    Sentence Representation Size $k^h$& 230\\
    Word Dimension $k^w$& 50\\
    Position Dimension $k^p$&5\\
    Pre-train Learning Rate $\lambda_P$&0.005\\
    Fine-tune Learning Rate $\lambda_F$&0.002\\
    Dropout Probability $p$&0.5\\
    \hline
    \end{tabular}
    \caption{Parameter settings in \reds.}
    \label{tab:exp}
\end{table}
We preprocess the WikiTable corpus with PySpark to build index for anchor entity pairs. On a single machine with two 8-core E5 CPUs and 256 GB memory, this processing takes around 20 minutes.

We use word embeddings from \citep{lin2016neural} for initialization, which are learned by word2vec tool\footnote{https://code.google.com/archive/p/word2vec/} on NYT corpus. The vocabulary is composed of words that appear more than 100 times in the corpus and words in an entity mention are concatenated as a single word.

To see the effect of 2-hop DS more directly, we set most parameters in \reds\ following \citet{lin2016neural}. Since the original NYT dataset only contains training and testing set, we randomly sample \nop{separate}20\% training data for development. We first pre-train a PCNN+ATT model with only $S$ and sentence-level selective attention. This BASE model converges in around 100 epochs. We then fine-tune the entire model with $S^T$ and bag-level aggregation added, which can finish within 50 epochs. Some key parameter settings in \reds\ are summarized in Table \ref{tab:exp}. %used in our experiments 

In testing phase, inference using 2-hop DS is slower, because the average size of $S^T$ is about 100 times that of $S$. With single 2080ti GPU, one full pass of testing data takes around 37s using \reds, compared with 12s using BASE model.

\subsection{Results}
\begin{figure}[t!]
\centering
\includegraphics[width=0.9\linewidth]{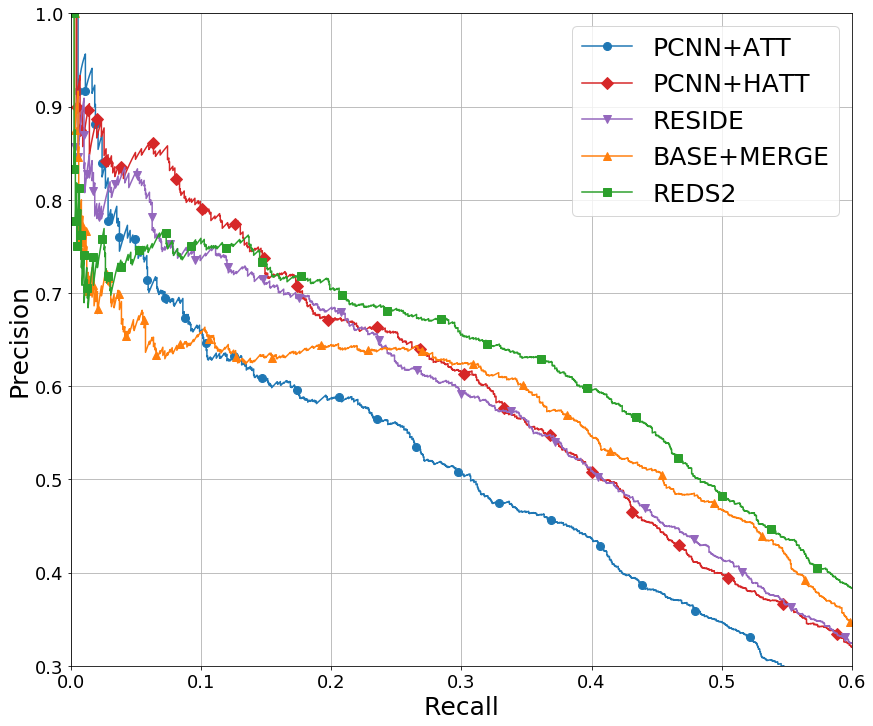}
\caption{Precision-recall curves for the proposed model and various baselines. }
\label{fig:overall}
\end{figure}
\begin{table}[t!]
    \centering
    \resizebox{\linewidth}{!}{
    \begin{tabular}{c|cccc}
    \hline
         Method\nop{Metric}& P@0.1&P@0.2&P@0.3&AUC \\
         \hline
         PCNN+ATT&65.9&58.7&50.4&35.8\\
         PCNN+HATT&\textbf{79.3}&67.2&61.6&42.0\\
         RESIDE&73.6&68.4&59.5&41.5\\
         \hline \hline
         BASE+MERGE&65.9&64.5&62.4&41.2\\
         \textbf{\reds}&75.9&\textbf{70.4}&\textbf{65.5}&\textbf{44.7}\\
         \hline
    \end{tabular}}
    \caption{Comparison on Precision@recall and AUC. \nop{for the proposed model and various baselines (\%)}}
    \label{tab:overall}
\end{table}
\subsubsection{Overall Evaluation Results}
Evaluation results on all target entity pairs in testing set are shown in Figure \ref{fig:overall} and Table \ref{tab:overall}, from which we make the following observations: 

(1) Figure \ref{fig:overall} shows all models obtain a reasonable precision when recall is smaller than 0.05. With the recall gradually increasing, the performance of models with 2-hop DS drops slower than those existing methods without. From Figure \ref{tab:overall}, we can see simply merging $S^T$ with $S$ in BASE+MERGE can boost the performance of basic PCNN+ATT model, and even achieves higher precision than state-of-the-art models like PCNN+HATT when recall is greater than 0.3\nop{ and better AUC}. \nop{This demonstrates that the extra information obtained by 2-hop DS is helpful for relation extraction, especially when directly associated sentences are noisy} This demonstrates that models utilizing 2-hop DS are more robust and remain a reasonable precision when including more lower-ranked relational facts which tend to be more challenging to predict because of insufficient evidence. %fall in a high recall region which the model is\fromH{Why does this correspond to higher recalls?}.

(2) As shown in both Figure \ref{fig:overall} and Table \ref{tab:overall}, \reds\ achieves the best results among all the models. Even when compared with PCNN+HATT and RESIDE which adopt extra relation hierarchy and side information from KB, our model still enjoys a significant performance gain. This is because\nop{indicates that, compared with the conventional 1-hop DS methods,} our method can take advantage of the rich entity pair correlations in Web tables and leverage the extra information brought by 2-hop DS. We anticipate our \reds\ model can be further improved by using more advanced sentence encoders and extra mechanisms like reinforcement learning \cite{feng2018reinforcement} and adversarial training \cite{wu2017adversarial}, which we leave for future work.

\begin{table*}[t!]
\centering
\resizebox{\textwidth}{!}{
    \begin{tabular}{c|P{0.9cm}P{0.9cm}P{0.9cm}P{0.9cm}|P{0.9cm}P{0.9cm}P{0.9cm}P{0.9cm}|P{0.9cm}P{0.9cm}P{0.9cm}P{0.9cm}|P{0.9cm}P{0.9cm}P{0.9cm}P{0.9cm}}
    \hline
    \multicolumn{1}{c|}{\multirow{2}{*}{Test Mode}} & \multicolumn{4}{c|}{\multirow{2}{*}{SINGLE}} & \multicolumn{12}{c}{MULTIPLE}                                                \\
\multicolumn{1}{c|}{}                           & \multicolumn{4}{c|}{}                        & \multicolumn{4}{c}{ONE} & \multicolumn{4}{c}{TWO} & \multicolumn{4}{c}{ALL} \\
\hline
 Metric & P@0.1 &P@0.2& P@0.3 &AUC& P@0.1 &P@0.2& P@0.3 &AUC& P@0.1 &P@0.2& P@0.3&AUC&P@0.1 &P@0.2& P@0.3 &AUC\\
    \hline
    PCNN+ATT&57.3&53.0&41.3&30.0&72.9&66.4&57.5&39.0&75.3&69.7&63.8&45.9&80.5&68.6&64.2&48.6\\
    RESIDE&66.9&60.5&51.8&36.2&79.7&70.4&57.4&42.1&77.7&76.1&67.6&46.5&83.2&\textbf{80.3}&73.2&51.9\\
    PCNN+HATT&\textbf{70.1}&59.1&50.8&34.8&74.5&67.4&57.2&40.1&78.4&68.9&65.9&44.5&\textbf{86.4}&75.5&70.0&49.7\\
    \hline \hline
    BASE+MERGE&62.2&58.5&55.2&35.6&71.4&74.9&73.2&51.3&70.0&71.8&73.9&52.0&70.7&71.1&73.2&52.0\\
    \textbf{\reds}&69.1&\textbf{61.1}&\textbf{57.4}&\textbf{37.5}&\textbf{82.4}&\textbf{81.9}&\textbf{78.1}&\textbf{56.3}&\textbf{81.4}&\textbf{79.6}&\textbf{76.9}&\textbf{57.2}&82.4&79.6&\textbf{76.6}&\textbf{57.6}\\
    \hline 
    \end{tabular}}
    \caption{Comparison on Precision@recall and AUC under different testing settings, detailed in Section \ref{sec:sentence}.}
    \label{tab:sentence_num}
\end{table*}

\begin{figure}
\centering
\includegraphics[width=0.9\linewidth]{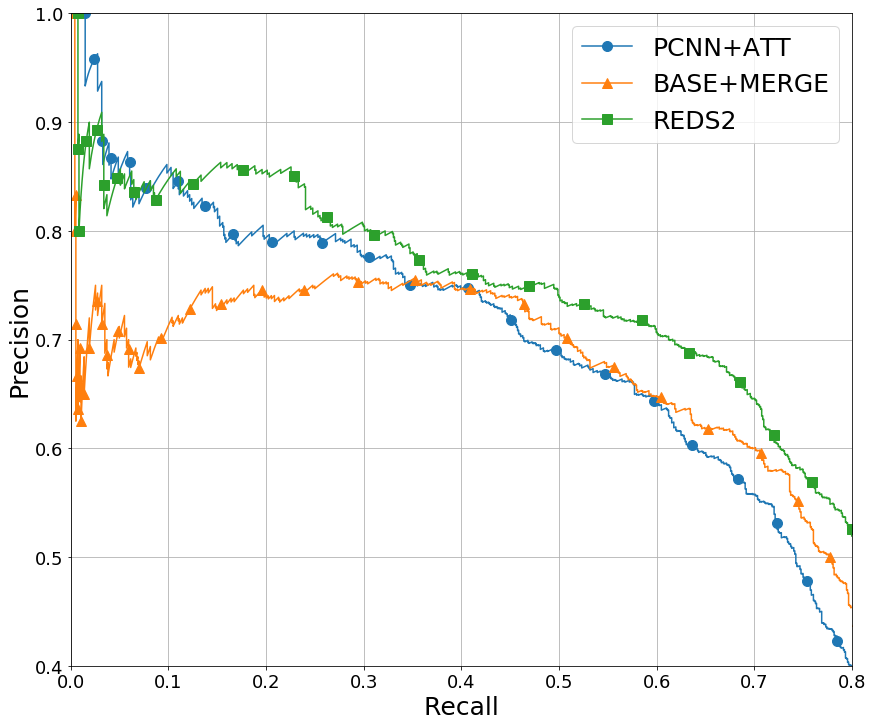}
\caption{Precision-recall curves on the subset of test entity pairs whose $S^T$ is not empty, to better show the effect of hierarchical bag aggregation design.\nop{Precision-recall curves for the proposed model and several baselines which also uses 2-hop DS from tables. This is evaluated on the subset of test that $S^T$ is not empty.}}
\label{fig:table_agg}
\end{figure}

%\subsection{Effect of Dynamic Bag Aggregation}
\begin{table}[]
    \centering
    \begin{tabular}{c|cccc}
    \hline
    max $|S^T|$ & P@0.1&P@0.2&P@0.3&AUC \\
    \hline
     10&57.9&55.8&51.5&36.2\\
     50&69.4&65.7&60.8&42.2\\
     100&70.4&66.7&62.3&43.2\\
     200&72.8&68.4&63.4&44.0\\
     300&\textbf{75.9}&\textbf{70.4}&\textbf{65.5}&\textbf{44.7}\\
     \hline
    \end{tabular}
    \caption{Effect of the table expanded sentence bag size $|S^T|$ on Precision@recall and AUC.\nop{Quantitative Results for the proposed model with different number of table expanded sentences sampled.}}
    \label{tab:table_bag_size}
\end{table}
\subsubsection{Effect of Hierarchical Bag Aggregation}
To further show the effect of our hierarchical bag aggregation design, here we also plot precision-recall curves in Figure \ref{fig:table_agg} on a subset of entity pairs in the test set (i.e., 4832 in total according to Table \ref{tab:datastat}) whose table expanded sentence bag $S^T$ is not empty. 

One main challenge of using 2-hop DS is it brings more noise. As shown in Table \ref{tab:datastat}, for entity pair with nonempty $S^T$, the size of $S^T$ is usually tens of times the size of $S$. From Figure \ref{fig:table_agg} we can see BASE+MERGE performs much worse compared with PCNN+ATT when recall is smaller than 0.2. {This is because \tbag\ tends to be much larger than \bag, and the model has a larger chance to attend to the noisy sentences obtained from 2-hop DS\nop{after reweighting the merged set with selective attention}. While ignoring the information in its directly associated sentences.}\nop{\xd{From Xiang: Because \tbag\ is much larger than \bag, thus have more influence. Is "biased" to "focus" ok}} We alleviate this problem by introducing hierarchical structure to first aggregate the two sets separately and then weight and sum them together. The proposed \reds\ model has a comparable precision with PCNN+ATT in the beginning and gradually outperform it.

%\subsection{Effect of Sentence Number}\label{sec:sentence}
\subsubsection{Effect of Sentence Number}\label{sec:sentence}
\textbf{Number of sentences from 1-hop DS}. In the originally testing set, there are 79176 entity pairs that are associated with only one sentence, out of which 1149 actually have relations. We hope our model can improve performance on these long-tail entities. Following \citet{lin2016neural}, we design the following test settings to evaluate the effect of sentence number: the ``SINGLE'' test setting contains all entity pairs that correspond to only one sentence; the ``MULTIPLE'' test setting contains the rest of entity pairs that have at least two sentences associated. We further construct the ``ONE'' testing setting where we randomly select one sentence for each entity pair; the ``TWO'' setting where we randomly select two sentences for each entity pair and the ``ALL'' setting where we use all the associated sentences from MULTIPLE. We use all sentences in $S^T$ for each entity pair if it is nonempty. Results are shown in Table \ref{tab:sentence_num}, from which we can see that \reds\ and BASE+MERGE have 25.0\% and 18.7\% improvements under AUC compared with PCNN+ATT in the SINGLE setting. Although the performance of all models generally improves as the sentence number increases in MULTIPLE setting, models leveraging 2-hop DS are more stable and have smaller changes. These observations indicate that 2-hop DS is helpful when information obtained by 1-hop DS is insufficient.

\noindent\textbf{Number of sentences from 2-hop DS}. We also evaluate how the number of sentences obtained by 2-hop DS will affect the performance of our proposed model. In Table \ref{tab:table_bag_size}, we show the performance of \reds\ with different numbers of sentences sampled from $S^T$. We observe that: (1) Performance of \reds\ improves as the number of sentences sampled increases. This shows that the selective attention over $S^T$ can effectively take advantage of the extra information from 2-hop DS while filtering out noisy sentences. (2) Even with 50 randomly sampled sentences, our model \reds\ still has a higher AUC than all baselines in Table \ref{tab:overall}. This indicates information obtained by 2-hop DS is redundant, even a small portion can be beneficial to relation extraction. How to sample a representative set effectively is worth further exploring in future work. 

\begin{table}[]
    \centering
    \begin{tabular}{c|p{6cm}}
    \hline
    \nop{\multicolumn{2}{c}{Relation: \texttt{person.place\_of\_death}}\\
    \hline
    1-hop & ... who shimmied at locations like the rock and roll hall of fame in cleveland and graceland , the \textbf{elvis\_presley} mansion in \textbf{memphis}.
    \\
    \hline
    2-hop & ...but we have \textbf{mao\_zedong}'s shrine in the middle of \textbf{beijing}, which is ...\\
    \hline}
    \multicolumn{2}{c}{Relation: \texttt{country.capital}}\\
    \hline
    1-hop & ... the golden gate bridge and the petronas towers in \textbf{kuala\_lumpur}, \textbf{malaysia}, was experienced ...
    \\
    \hline
    2-hop & a friend from \textbf{cardiff} , the capital city of \textbf{wales} , lives for complex ...\\
    \hline
    \end{tabular}
    \caption{An example for case study, where the sentence with the highest attention weight is selected respectively from 1-hop and 2-hop sentence bag.}
    \label{tab:case}
\end{table}

\subsection{\nop{Evaluation of New Knowledge Discovery from Table Corpus}RE for Entity Pairs with Empty 1-hop Sentence Bag}
We observe that there are large amounts of entity pairs in the table corpus that have no associated sentences but have anchor entity pairs mentioned in the text corpus. By leveraging 2-hop distant supervision, we can do relation extraction for this set of entity pairs\nop{, and discover new knowledge}.\\
We extract a total number of 251917 entity pairs from the WikiTable dataset which do not exist in the NYT dataset but have at least one anchor entity pair that appear in the original NYT training data. We randomly sample 10000 examples and evaluate our trained model on them. Surprisingly, the relation extraction result is even better than the result on the NYT test data in Table \ref{tab:overall}, with an overall AUC of 54.7 and a P@0.3 of 71.1. This can be explained partly by two observations\nop{This comes from two advantages of using table corpus and 2-hop DS}: (1) The table corpus generates higher-quality entity pairs, 18\% of extracted entity pairs have non-NA relations, compared with only 1.8\% in NYT test data. (2) The newly extracted entity pairs have 14 useful anchor entity pairs and 175 2-hop DS sentences on average, which give ample information for prediction. \nop{We think such new extracted knowledge will benefit many downstream tasks.}This study shows that for two entities that have no directly associated sentences, it is possible to utilize the 2-hop DS to predict their relations accurately.

\subsection{Case Study and Error Analysis}
In addition to the motivating example from the training set shown in Figure \ref{fig:table_example}, we also demonstrate how 2-hop DS helped relation extraction using an example from the testing set in Table \ref{tab:case}.\nop{For \bag\ and \tbag, we show the corresponding sentence with highest attention weight.} As we can see, the sentence with the highest attention weight in \bag\ does not express the desired relation between the target entity pair whereas that in 2-hop DS bag clearly indicates the \texttt{country.capital} relation.

\nop{We also analyze some cases where \reds\ makes mistakes and gives worse predictions. We select 50 most divisive examples between BASE and \reds\ and categorize the errors.}We also conduct an error analysis by analyzing examples where REDS2 gives worse predictions than BASE (e.g., assigns a lower score to a correct relation or a higher score to a wrong relation), and 50 examples with most disparity in the two methods' scores are selected. We find that 29 examples have wrong labels caused by KB incompleteness and our model in fact makes the right prediction. 11 examples are due to errors in column processing (e.g., errors in NE/subject column selection and entity linking), 9 are caused by anchor entity pairs with differet relations (e.g., \nop{\textbf{(Connecticut, Boston)} and \textbf{(Missouri, St.\_Louis)} are in the same table \textit{Historical capitals in the United States of America}, but only the latter has relation \texttt{location.contains}}\textbf{(Greece, Atlanta)} and \textbf{(Mexico, Xalapa)} are in the same table \textit{``National Records in High Jump"} under columns \textbf{(Nation, Place)}, but only the latter has relation \texttt{location.contains}), and 1 is because of wrong information in the original table. %mistake  \fromH{just curious, what mistake?} %\xd{\url{https://en.wikipedia.org/w/index.php?title=List_of_Avril_Lavigne_concert_tours&oldid=779544118} in this table, tropicana\_field is in tampa, which is wrong. This is corrected in Wikipedia now}

\section{Conclusion and Future Work}
This paper introduces 2-hop distant supervision for relation extraction, based on the intuition that entity pairs in relational Web tables often share common relations. Given a target entity pair, we define and find its anchor entity pairs via Web tables and collect all sentences that mention the anchor entity pairs to help relation prediction. We develop a new neural RE method \reds\ in the multi-instance learning paradigm which fuses information from 1-hop DS and 2-hop DS using a hierarchical model structure, and substantially outperforms existing RE methods on a benchmark dataset. Interesting future work includes: (1) Given that information from 2-hop DS is redundant and noisy, we can explore smarter sampling and/or better bag-level aggregation methods to capture the most representative information. (2) Metadata in Web tables like headers and column names also contain rich information, which can be incorporated to further improve RE performance. \nop{(3) Current NYT dataset is very imbalanced. Only 5\% of entity pairs have non-NA relation, and focus on a few relation types. On the other hand, we observe that entity pairs extracted from tables usually have meaningful and diverse relations. We can create a new high-quality relation extraction dataset by combining web-scale text and table corpus.}

\section{Acknowledgments}
This research was sponsored in part by the Army Research Office under cooperative agreements NSF Grant
IIS1815674, W911NF-17-1-0412, Fujitsu gift grant, and Ohio Supercomputer Center [8]. The views and conclusions contained herein are those of the authors and should not be interpreted as representing the official policies, either expressed or implied, of the Army Research Office or the U.S. Government. The U.S. Government is authorized to reproduce and distribute reprints for Government purposes notwithstanding any copyright notice herein.

%show that \reds\ can {consistently outperform various baselines across different settings by a substantial margin.}\footnote{Our source code and datasets are at \url{anonymous.com}.} \nop{We also conduct case studies to show that our model is capable of finding new knowledge by incorporating Web tables.}

\bibliography{ref}
\bibliographystyle{acl_natbib}
\end{document}